\documentclass[letterpaper]{article} 
\usepackage{aaai2026}  
\usepackage{times}  
\usepackage{helvet}  
\usepackage{courier}  
\usepackage[hyphens]{url}  
\usepackage{graphicx} 
\usepackage{natbib}  
\usepackage{caption} 
\usepackage{algorithm}
\usepackage{algorithmic}

\usepackage{amsmath}
\usepackage{amsthm}
\usepackage{amssymb}
\usepackage{mathrsfs}
\usepackage{bm}

\usepackage[table]{xcolor}

\usepackage{xcolor}
\usepackage{soul}
\definecolor{WarnPink}{rgb}{0.9176, 0.7215, 0.7215}
\definecolor{GoodGreen}{rgb}{0.5019, 0.9215, 0.6039}
\newcommand{\warn}[1]{\sethlcolor{WarnPink}\hl{#1}}

\usepackage{multirow}
\usepackage{booktabs}
\usepackage{bbding}

\usepackage{newfloat}
\usepackage{listings}
\DeclareCaptionStyle{ruled}{labelfont=normalfont,labelsep=colon,strut=off} 
\floatstyle{ruled}
\newfloat{listing}{tb}{lst}{}
\floatname{listing}{Listing}
\title{Mass Concept Erasure in Diffusion Models with Concept Hierarchy}
\author{
    Jiahang Tu\textsuperscript{\rm 1},
    Ye Li\textsuperscript{\rm 1},
    Yiming Wu\textsuperscript{\rm 2},
    Hanbin Zhao\thanks{Corresponding author.}\textsuperscript{\rm 1, \rm3},
    Chao Zhang\textsuperscript{\rm 1},
    Hui Qian\textsuperscript{\rm 1}
}

\affiliations{
    \textsuperscript{\rm 1}College of Computer Science and Technology, Zhejiang University\\
    \textsuperscript{\rm 2}School of Computing and Data Science, The University of Hong Kong\\
    \textsuperscript{\rm 3}Zhejiang Key Laboratory of Intelligent High Speed UAV Technology\\
    \{tujiahang, yeli1, zhaohanbin, zczju, qianhui\}@zju.edu.cn, yimingwu0@gmail.com
}

\usepackage{bibentry}

\begin{document}

\maketitle

\begin{abstract}
The success of diffusion models has raised concerns about the generation of unsafe or harmful content, prompting concept erasure approaches that fine-tune modules to suppress specific concepts while preserving general generative capabilities. However, as the number of erased concepts grows, these methods often become inefficient and ineffective, since each concept requires a separate set of fine-tuned parameters and may degrade the overall generation quality. In this work, we propose a supertype-subtype concept hierarchy that organizes erased concepts into a parent–child structure. Each erased concept is treated as a child node, and semantically related concepts (\emph{e.g.}, macaw, and bald eagle) are grouped under a shared parent node, referred to as a supertype concept (\emph{e.g.}, bird). Rather than erasing concepts individually, we introduce an effective and efficient group-wise suppression method, where semantically similar concepts are grouped and erased jointly by sharing a single set of learnable parameters. During the erasure phase, standard diffusion regularization is applied to preserve denoising process in unmasked regions. To mitigate the degradation of supertype generation caused by excessive erasure of semantically related subtypes, we propose a novel method called \textbf{Su}pertype-\textbf{P}reserving \textbf{Lo}w-\textbf{R}ank \textbf{A}daptation (SuPLoRA), which encodes the supertype concept information in the frozen down-projection matrix and updates only the up-projection matrix during erasure. Theoretical analysis demonstrates the effectiveness of SuPLoRA in mitigating generation performance degradation. We construct a more challenging benchmark that requires simultaneous erasure of concepts across diverse domains, including celebrities, objects, and pornographic content. Comprehensive experiments demonstrate that our method achieves a superior balance between effective multi-concept erasure and the preservation of desirable generative performance.
\end{abstract}

\begin{links}
    \link{Code}{https://github.com/TtuHamg/SuPLoRA}
\end{links}

\section{Introduction}
Recent advances in diffusion models \cite{song2020denoising, ho2020denoising, tu2025driveditfit} have greatly improved text-to-image (T2I) generation, enabling users to produce high-quality and realistic images from simple text prompts. Tools like Stable Diffusion (SD) \cite{Rombach_2022_CVPR, podell2023sdxl}, MidJourney \cite{midjourney}, and Flux \cite{flux2024} highlight this capability. However, these advances raise major ethics \cite{jiang2023ai,wang2024gad}, privacy \cite{mirsky2021creation}, and safety concerns \cite{wang2025efficiently}, as models often learn undesirable concepts, such as copyrighted materials, offensive content, and sensitive personal information, from unfiltered datasets \cite{Rombach_2022_CVPR}. Even with the high-cost data cleaning, diffusion models can still learn and generate unsafe content on filtered datasets \cite{schramowski2023safe}. 

To tackle this problem, various concept erasure methods have been proposed to \textbf{suppress the generation of concepts to be erased} while \textbf{preserving the model’s capacity to generate general ones}. Early studies \cite{gandikota2023erasing, schramowski2023safe, heng2023selective, fan2023salun, li2024safegen, yoon2024safree, tu2025sdwv, feng2025fg, li2025sculpting, feng2024lw2g, feng2025pectp} tune specific diffusion modules to erase single concepts. Recently, growing efforts have focused on mass concept erasure \cite{kumari2023ablating, zhang2024forget, zhao2024separable, lu2024mace, gandikota2024unified, lyu2024one}, often via adapter tuning \cite{houlsby2019parameter} or low-rank adaptation (LoRA) \cite{hu2022lora}. These methods typically inject learnable parameters per concept and fine-tune them to suppress their generation.

\begin{figure}[tb]
  \centering
  \includegraphics[width=1.0\linewidth]{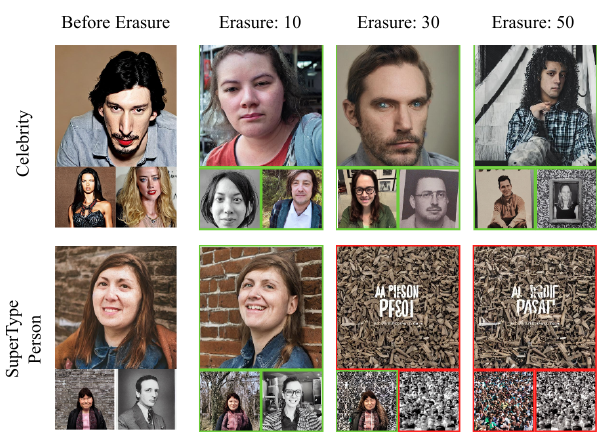}
  \vspace{-1.5em}
  \caption{As more celebrities are erased, Stable Diffusion shows clear degradation in generating the supertype \emph{person}. Results are shown for three representative celebrities (“\emph{Adam Driver}”, “\emph{Adriana Lima}”, “\emph{Amber Heard}”). Green boxes in the first row indicate successful erasure. In the second row, green boxes indicate the generation capability of the supertype is preserved, while red boxes indicate failure.}
  \label{fig:collapse}
\end{figure}

However, these concept-wise erasure methods face two primary issues. Firstly, since the number of fine-tuned parameter sets grows linearly with the number of concepts to be erased, concept-wise erasure methods become inefficient as the number of erased concepts increases. This significantly increases storage overhead, making these methods impractical for real-world applications that require the erasure of a large number of concepts \cite{gpt2023}. Secondly, since erasing a concept requires the model to suppress its learned generation patterns \cite{schramowski2023safe,gandikota2023erasing,cfg,meicosketcher}, repeatedly applying this process across multiple concepts inevitably degrades the model's generative capacity on general concepts \cite{lu2024mace,zhao2024separable}. For instance, in the celebrity erasure task, celebrities have distinct appearances, but they all belong to the supertype \cite{dai2023hierarchical, wang2025unleashing} ``person''. When the model is instructed to forget an increasing number of celebrity identities, it will inevitably suppress visual features that are not only specific to those individuals but also essential for representing the supertype ``person''. As illustrated in Fig.\ref{fig:collapse}, increasing the number of erased celebrities results in noticeable degradation in the model's ability to generate the supertype ``person''.

To address these issues, we leverage the semantic relationship among concepts to be erased and construct a concept hierarchy \cite{dai2023hierarchical, an2025concept} by organizing them into a parent–child structure. In this hierarchy, each erased concept is treated as a child node, and semantically related concepts, such as different species of birds (\emph{e.g.}, jay, macaw, bald eagle), are grouped under a common parent node, which we define as a supertype concept (\emph{e.g.}, bird). Instead of erasing concepts individually in previous methods, our supertype-subtype hierarchy enables group-wise erasure, where grouped child concepts are erased jointly in a more efficient manner. Specifically, we adopt MACE's \cite{lu2024mace} attention-based suppression to reduce the model focus on regions associated with the grouped concepts, and apply standard diffusion regularization to unmasked regions to preserve the model’s denoising capability. To retain the generation capability for supertype concepts, we propose a \textbf{Su}pertype-\textbf{P}reserving \textbf{Lo}w-\textbf{R}ank \textbf{A}daptation (SuPLoRA), which designs the bases for representing the supertype concept subspace in the embedding space, initializes the down-projection matrix in SuPLoRA using orthogonal bases, and only trains the up-projection matrix. We further theoretically analyze the relationship between the update of down-projection matrix and the parameter of up-projection matrix in erasure setting, and prove that SuPLoRA mitigates performance degradation of supertype concepts. To evaluate the scalability and robustness of our method, we construct a more challenging benchmark than previous studies, which typically focus on erasing concepts from a single category. Our benchmark involves simultaneous erasure of concepts across multiple domains, including celebrities, objects, and pornographic content. Experimental results show that our method achieves a more favorable balance between the erasure of undesired concepts and the preservation of generative quality.

Our contributions are summarized as follows: 
\begin{itemize}
    \item We propose a concept hierarchy that organizes erased concepts into parent–child relationships. Based on this structure, we introduce a group-wise erasure strategy that jointly erases semantically related concepts under a shared supertype concept, improving performance efficiency over traditional concept-wise methods.
    \item We design SuPLoRA, a \textbf{Su}pertype-\textbf{P}reserving \textbf{Lo}w-\textbf{R}ank \textbf{A}daptation, to tackle the generation degradation of supertype concepts. Theoretical analysis ensures the effectiveness of SuPLoRA in mitigating generation performance degradation.
    \item We construct a more challenging benchmark spanning diverse concept, and demonstrate that our method achieves a superior trade-off between mass concept erasure and the preservation of generative quality.
\end{itemize}

\section{Related Work}

\subsection{Inference-Time Intervention}
Inference-time intervention methods aim to block undesired content by modifying the sampling process without changing model parameters. A common strategy \cite{schramowski2023safe, negative_prompt} is to adjust classifier-free guidance \cite{cfg, wang2024belm}, as in Safe Latent Diffusion \cite{schramowski2023safe}, which steers latent representations away from erased concepts. Another line of research \cite{tu2025sdwv, yoon2024safree, li2024get, wang2024precise} focuses on manipulating the text embeddings used for conditioning. CE-SDWV \cite{tu2025sdwv} constructs a semantic space representing the erased concepts and dynamically suppresses the corresponding semantic information that is hidden in the text embeddings. SAFREE \cite{yoon2024safree} applies subspace projection and adaptive re-attention to eliminate unsafe semantic directions in CLIP embeddings. While such methods avoid model fine-tuning and offer efficient erasure, they are fragile in practice, as the intervention can be bypassed simply by disabling the module during inference \cite{rando2022red}.

\subsection{Tuning-based Erasure}
To support safer model release, existing studies have explored fine-tuning methods to erase targeted concepts from pre-trained models. These approaches can be broadly categorized by the number of concepts they handle. Several works focus on single concept erasure \cite{gandikota2023erasing, heng2023selective, zhang2024forget, fan2023salun, li2024safegen, gao2025eraseanything, li2025set}, aiming to remove one concept at a time. ESD \cite{gandikota2023erasing} and AC \cite{kumari2023ablating} align erased concepts with supertypes (\emph{e.g.}, “grumpy cat” → “cat”), by fine-tuning cross-attention layers. SalUN \cite{fan2023salun} adopts a saliency-driven strategy that selectively updates parameters most associated with the target concept, minimizing side effects on general generation. In contrast, mass concept erasure \cite{zhao2024separable, lu2024mace, gandikota2024unified, lyu2024one, kumari2023ablating, huang2023receler, li2025sculpting} targets multiple concepts simultaneously. For example, ConceptPrune \cite{chavhan2024conceptprune} prunes a union of “expert neurons” responsive to each concept, achieving erasure of ten object classes. MACE \cite{lu2024mace} leverages the LoRA fine-tuning technique, where each LoRA module is trained to erase a specific concept. However, most existing methods follow a concept-wise paradigm, in which each concept requires its own fine-tuned module. As the number of concepts increases, the total number of trainable parameters grows linearly, resulting in substantial storage overhead. This severely limits the scalability of these methods in real-world scenarios, where a large number of concepts may need to be erased simultaneously.

\subsection{Preservation on Unerased Concepts}
While existing methods effectively remove specific concepts, preserving generation for unerased concepts, particularly in mass concept erasure scenarios, remains a critical and underexplored challenge. To mitigate this, Selective Amnesia (SA) \cite{heng2023selective} introduces a regularization term inspired by lifelong learning principles \cite{wang2022learning}, encouraging the model to retain knowledge of unerased content. UCE \cite{gandikota2024unified} extends the TIME framework \cite{orgad2023editing} by applying an auxiliary loss on a set of predefined preserved concepts and deriving a closed-form solution that balances erasure and retention objectives. SPM \cite{lyu2024one} proposes an anchoring loss to protect distant, unrelated concepts during sequential erasure. Despite these efforts, a key limitation persists: as more concepts are erased, the model’s ability to generate their shared supertype concept often degrades. In this work, we explicitly identify and address this issue by proposing SuPLoRA, a principled approach for preserving generation of unerased concepts.

\begin{figure*}[t]
  \centering
  \includegraphics[width=1.0\linewidth]{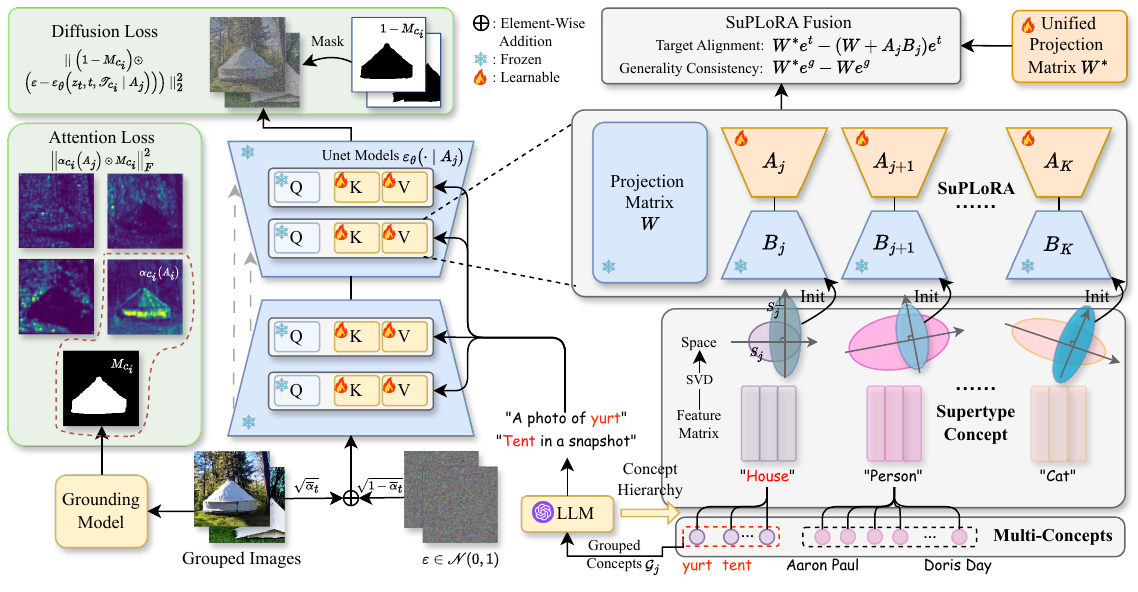}
  \caption{An overview of our mass concept erasure framework. We first construct a supertype–subtype concept hierarchy using LLMs (Sec.\ref{sec:hierarchy}). Leveraging this hierarchy, we introduce a group-wise suppression method that jointly erases grouped concepts via MACE’s attention modulation, and apply diffusion regularization to preserve the model’s denoising capability (Sec.\ref{sec:suppression}). To mitigate the degradation of supertype generation caused by excessive erasure of semantically related subtypes, we propose SuPLoRA, a supertype-preserving low-rank adaptation that freezes the down-projection matrix $\bm{B}_j$ (orthogonal to the supertype subspace) and updates only the up-projection matrix $\bm{A}_j$, as theoretically analyzed in Sec.\ref{sec:SuPLoRA}. Finally, we merge multiple SuPLoRA modules into a unified representation while preserving general generation ability (Sec.~\ref{sec:SuPLoRA}).}
  \label{fig:architecture}
\end{figure*}

\section{Methodology}
The objective of mass concept erasure is to remove a set of various concepts, while preserving the generative model's capability to produce high-fidelity outputs for general concepts.  Let $\mathcal{M}$ be a pre-trained generative model with zero-shot capabilities over a broad set of general concepts $\mathcal{C}^g$. Let $\mathcal{C}^t = \{ c^t_1, c^t_2, \dots, c^t_N \}$ denote the set of concepts targeted for erasure, where $C^t\cap C^g=\emptyset$. The goal is to suppress the model's ability to generate content corresponding to each $c^t_i \in \mathcal{C}$, without degrading its generation quality on the general concepts. As the number of erased concepts increases, special attention must be given to preserving the generation capabilities of supertype concepts $\mathcal{C}^p = \{ c^p_1, c^p_2, \dots, c^p_K \}$. The erasure algorithm $\mathscr{E}$ aims to produce a modified model $\mathcal{M}^{\prime}$ = $\mathscr{F}(\mathcal{M},\mathcal{C}^t, \mathcal{C}^g, \mathcal{C}^p )$ that modifies the mapping behavior as follows:
\begin{equation}
\begin{cases}
f_{\mathcal{M}^{\prime}}(\mathcal{T}_{\mathcal{C}^t}) \not\stackrel{}{\longrightarrow} \mathcal{I}_{\mathcal{C}^t}, \\
f_{\mathcal{M}^{\prime}}(\mathcal{T}_{\mathcal{C}^g}) \stackrel{}{\longrightarrow} \mathcal{I}_{\mathcal{C}^g},
f_{\mathcal{M}^{\prime}}(\mathcal{T}_{\mathcal{C}^p}) \stackrel{}{\longrightarrow} \mathcal{I}_{\mathcal{C}^p}.
\end{cases}
\end{equation}
Here, $f_{\mathcal{M}^{\prime}}(\mathcal{T})$ denotes the mapping from textual descriptions of concepts to image outputs $\mathcal{I}$, and $\not\stackrel{}{\longrightarrow}$ indicates that the original mapping no longer holds after erasure.


\subsection{Concept Hierarchy Construction}
\label{sec:hierarchy}
Previous approaches to mass concept erasure typically allocate a separate set of fine-tuning parameters for each target concept, treating each concept independently. However, many erased concepts exhibit semantic similarity, and overlooking these relationships leads to redundant parameter usage and storage overhead. To address this limitation, we propose to construct a concept hierarchy by exploiting the inherent semantic structure among concepts.

In our hierarchy, each concept to be erased is represented as a child node, while semantically related concepts are grouped under a common parent node, which we refer to as a supertype concept. We define the relationship as:
\begin{align}
    \mathcal{G}_j = \{ c^t_i \in \mathcal{C}^t \mid g(c^t_i) = c^p_j \},
\end{align}
where $g:\mathcal{C}^g\rightarrow \mathcal{C}^p$ s a mapping function that assigns each erased concept $c^t_i$ to its corresponding supertype $c^p_j$. For example, as shown in Fig.\ref{fig:architecture}, different celebrities such as ``Aaron Paul'' and ``Doris Day'' are different erased concepts, but they share a common supertype ``person''. In this work, we construct a two-level concept hierarchy based on the supertype-subtype relationship among concepts. This hierarchical structure allows us to erase semantically similar concepts jointly by learning a shared set of parameters associated with their supertype. To build this hierarchy, we leverage the advanced semantic understanding of large language models \cite{gpt2023} (LLMs). Specifically, based on the semantic similarity among different concepts, we cluster multiple erased concepts into groups such that concepts within each group share a high-level semantic abstraction. Each group is then associated with a single supertype concept, identified using an LLM. This concept hierarchy lays the foundation for the group-wise suppression method described in the next section. Details on how supertype concepts are derived using LLMs are provided in App. B.1, and experiments on constructing a more complex multi-level concept hierarchy are provided in App. C.4.

\subsection{Group-wise Suppression}
\label{sec:suppression}
To erase multiple concepts efficiently, we propose a group-wise suppression strategy based on the attention-based suppression method introduced in MACE \cite{lu2024mace}. Inspired by MACE, our method minimizes the model’s attention to concept-relevant regions to suppress undesired concepts. Unlike prior methods that treat each concept independently, we perform suppression at the supertype level, where semantically similar concepts are grouped and erased jointly by sharing a single set of learnable parameters. This design is motivated by our assumption that related concepts (\emph{e.g.}, jay, macaw, bald eagle) can be represented within a shared semantic abstraction (\emph{e.g.}, bird), and thus can be erased together through a unified parameter set. This reduces the total number of parameter sets from the number of erased concepts to the number of supertype groups, significantly improving parameter efficiency.

To achieve this, we utilize segmentation priors obtained from external grounding models \cite{liu2024grounding} to identify concept-relevant areas $\bm{M}_{c_i}$ in the image. During training, the attention scores of concept-related tokens are selectively discouraged from attending to these areas, thereby weakening their spatial grounding. The erasure objective is defined as follows:
\begin{align}
\mathcal{L}_{\text{attn}} = \mathbb{E}_{c_i\in \mathcal{G}_j, t,l} \left[ \Vert \bm{\alpha}_{c_i}^{t,l}(\bm{A}_j) \odot \bm{M}_{c_i} \Vert_F^2 \right], \label{eq:erasure}
\end{align}
where $\bm{\alpha}_{c_i}^{t,l}(\bm{A}_j)$ denotes the attention map of concept token $c_i$ at layer $l$ and timestep $t$, whose values are influenced by the learnable parameters $\bm{A}_j$ injected into the attention projection layers.

While attention suppression effectively erase multiple concepts, it may also impair the model’s ability to perform denoising, a core capability of diffusion-based generation. To address this, we apply standard diffusion training to the regions not associated with the erased concept:
\begin{align}
    \mathcal{L}_{\text{Diff}} = \mathbb{E}_{c_i\in \mathcal{G}_j,t,\bm{\epsilon} } & \left[ \Vert (1-\bm{M}_{c_i}) \odot(\bm{\epsilon} - \epsilon_\theta(\bm{z_t}, t, \mathcal{T}_{c_i}|\bm{A}_j)) \Vert_2^2 \right], \label{eq:diffusion}
\end{align}
where $\epsilon_\theta(\cdot|\bm{A}_j)$ denotes the UNet models with learnable parameter $\bm{A}_j$. By fine-tuning the parameter $\bm{A}_j$, we aim to minimize the final loss:
\begin{align}
    \mathcal{L} = \mathcal{L}_{\text{attn}} + \lambda\mathcal{L}_{\text{Diff}}.
\end{align}

\subsection{Supertype-Preserving Low-Rank Adaptation}
\label{sec:SuPLoRA}
Originally proposed as a parameter-efficient fine-tuning technique for large language models~\cite{gpt2023} (LLMs), LoRA \cite{hu2022lora} has recently gained increasing attention in the vision community. It operates by injecting a pair of low-rank matrices, named as a down-projection matrix $\bm{B}$ and a up-projection matrix $\bm{A}$, into the weight layers of pre-trained models. This design enables the model to acquire task-specific knowledge while keeping the original weights largely unchanged. In group-wise erasure, the forward pass of a linear layer can be formulated as follows:
\begin{align}
\label{eq:lora}
    \bm{o}_j = & \bm{W}\bm{h}_j+\bm{A}_j\bm{B}_j\bm{h}_j, \quad j\in[1,2,\dots,K],
\end{align}
where $K$ denotes the number of supertype concept. Here, $\bm{h}_j$ and $\bm{o}_j$ represent the input and output of the linear layer, respectively. In the following paragraph, we analyze how to preserve the generative capability of a supertype concept during the erasure process.

\emph{We start by analyzing the relationship between fine-tuning SuPLoRA and directly tuning pre-trained weight $\bm{W}$.} Suppose the $j^{th}$ grouped concepts are targeted for erasure. Concept removal can be achieved either by directly modifying the pre-trained weights $\bm{W}$ or by learning the LoRA parameters $\bm{A}_j$ and $\bm{B}_j$. 

If we opt to modify $\bm{W}$ directly, the gradient of the loss $\mathcal{L}$ with respect to $\bm{W}$ is computed via the chain rule:
\begin{align}
    \frac{\partial \mathcal{L}}{\partial \bm{W}} = & \frac{\partial \mathcal{L}}{\partial \bm{o}_j}\frac{\partial \bm{o}_j}{\partial \bm{W}}=\frac{\partial \mathcal{L}}{\partial \bm{o}_j}\bm{h}_j^{T}.
\end{align}
With a learning rate $\alpha$, the update to $\bm{W}$ is $\Delta\bm{W}=-\alpha\frac{\partial \mathcal{L}}{\partial \bm{W}}$. Thus, the corresponding change in the erased matrix $\bm{W}^\prime=\bm{W}+\bm{A}_j\bm{B}_j$ is:
\begin{align}
\label{eq:delta_W_prime}
    \Delta_{\bm{W}} \bm{W}^\prime = & [\bm{W}+\Delta\bm{W}+\bm{A}_{j}\bm{B}_{j}] -(\bm{W}+\bm{A}_{j}\bm{B}_{j})\nonumber \\
    = & \Delta \bm{W}=-\alpha\frac{\partial \mathcal{L}}{\partial \bm{W}}=-\alpha\frac{\partial \mathcal{L}}{\partial \bm{o}_j}\bm{h}_j^{T}.
\end{align}
On the other hand, if we fine-tune $\bm{A}_j$ while keeping $\bm{W}$ and $\bm{B}_j$ fixed in Eq.\ref{eq:lora}, its gradient is:
\begin{align}
\label{eq:grad_A_j}
        \frac{\partial \mathcal{L}}{\partial \bm{A}_{j}} = & \frac{\partial \mathcal{L}}{\partial \bm{o}_j}\frac{\partial \bm{o}_j}{\partial \bm{A}_{j}}=\frac{\partial \mathcal{L}}{\partial \bm{o}_j}\bm{h}_j^{T}\bm{B}_{j}^{T}.
\end{align}
Then, the update to $\bm{A}_j$ becomes $\Delta\bm{A}_j=-\alpha\frac{\partial \mathcal{L}}{\partial \bm{A}_j}$, and the change in the erased matrix $\bm{W^\prime}$ is:
\begin{align}
    \Delta_{\bm{A}_{j}} \bm{W}^\prime = & [\bm{W}+(\bm{A}_{j}+\Delta \bm{A}_{j})\bm{B}_{j}]-(\bm{W} + \bm{A}_{j}\bm{B}_{j})\nonumber\\
    =&\Delta \bm{A}_{j}\bm{B}_{j}=-\alpha\frac{\partial \mathcal{L}}{\partial \bm{A}_{j}}\bm{B}_{j} \nonumber \\
    =&\Delta_{\bm{W}} \bm{W}^\prime\bm{B}_{j}^{T}\bm{B}_{j}. \quad \text{(based on Eq.~\ref{eq:delta_W_prime})} \label{eq:projection}
\end{align}
In multi-domain concept erasure, Eq.\ref{eq:projection} reveals that \emph{\textbf{fine-tuning $\bm{A}_j$ of the $j^{th}$ grouped concepts is equivalent to modifying the pre-trained weight $\bm{W}$ within a subspace $\mathcal{S}_j^\perp$ defined by the projection matrix $\bm{B}_j^{T} \bm{B}_j$}}. A similar view has been adopted in continual learning settings \cite{liang2024inflora}. Building upon the above formulation, \emph{\textbf{if the subspace $\mathcal{S}_j^\perp$ defined by $\bm{B}_j$ is carefully designed to be orthogonal to the gradient subspace $\mathcal{S}_j$ associated with the learning of the supertype concept, then updates of $\bm{A}_j$ will lie in a direction orthogonal to that of the supertype concept gradients. Consequently, freezing $\bm{B}_j$ and fine-tuning $\bm{A}_j$ for concept erasure will not interfere with the model’s generation ability of the supertype concept.}} This intuition aligns with findings in prior work on gradient orthogonality for task interference mitigation \cite{saha2021gradient}.

\emph{We then describe how to construct the subspace $\mathcal{S}^\perp_j$ and initialize $\bm{B}_j$}. Prior studies \cite{liang2023adaptive, saha2021gradient} have demonstrated that the gradients of linear layers lie within the span of the input space. Leveraging this insight, SuPLoRA approximates the gradient subspace $\mathcal{S}_j$ for the $j^{th}$ supertype concept using the input matrix $\bm{H}_{S_j}$ of pre-trained weight $\bm{W}$. When SuPLoRA is fine-tuned in the key and value projections of the cross-attention layer, $\bm{H}_{S_j}$ corresponds to the embeddings of the supertype concept descriptions. Typically, singular value decomposition (SVD) is applied to $\bm{H}_{S_j}=\bm{U}_j\bm{\Sigma}_j\bm{V}^T_j$, and the subspace $\mathcal{S}_j=\text{span}\{\bm{u}_{1,j},\bm{u}_{2,j},\dots, \bm{u}_{r,j}\}$ is defined using the first $r$ principle components. The orthogonal complement of this subspace, $\mathcal{S}_j^\perp$, can be computed via the null space of $\mathcal{S}_j$ or the projection operation in \cite{liang2024inflora}. Consequently, SuPLoRA sets $\bm{B}_i$ to the basis of $\mathcal{S}_j^\perp$ and fine-tunes only $\bm{A}_j$ to erase the $j^{th}$ grouped concepts, while effectively preserving the generative capability of the supertype concept. 

\begin{table*}[t]
\resizebox{\textwidth}{!}{
\begin{tabular}{c|c|ccc|cc|cc|c|cc}
\toprule
\multicolumn{1}{c|}{\multirow{2}{*}{\textbf{Method}}} & \multicolumn{1}{c|}{\multirow{2}{*}{\textbf{Is Mass}}} & \multicolumn{3}{c|}{\textbf{Erasure Effect}}  & \multicolumn{2}{c|}{\textbf{Domain-Specific}} & \multicolumn{2}{c|}{\textbf{MS-COCO}}     & \textbf{Supertype} & \multicolumn{2}{c}{\textbf{Efficiency}} \\ 
\multicolumn{1}{c|}{}  & \multicolumn{1}{c|}{}   & \textbf{Cele Acc}($\downarrow$)   & \textbf{Obj Acc}($\downarrow$) & \textbf{NN}($\downarrow$) & \textbf{Cele Acc}($\uparrow$) & \textbf{Obj Acc}($\uparrow$) & \textbf{FID}($\downarrow$) & \textbf{CLIP Score}($\uparrow$) & \textbf{CLIP Score} ($\uparrow$) & \textbf{Storage/MB}($\downarrow$) & \textbf{Time/m}($\downarrow$) \\ \midrule
\rowcolor{gray!15}
\multicolumn{12}{l}{\textit{Methods that sacrifice generative performance for concept erasure}} \\
\rowcolor{gray!15}
ESD-x   & \XSolidBrush & 1.670\% & 15.40\% & 399 & \warn{3.375\%} & 52.50\% & 21.01 & 29.24 & 23.59 & 3379 & 2298 \\
\rowcolor{gray!15}
ESD-u   & \XSolidBrush & 0.000\% & 1.250\% & 59 & \warn{0.500\%} & \warn{7.625\%} & 34.59 & 25.21 & \warn{22.05} & 3379 & 2166 \\
\rowcolor{gray!15}
FMN  & \XSolidBrush & 0.000\% & 0.000\% & 0 & \warn{0.000\%} & \warn{0.000\%} & \warn{407.6} & \warn{16.85} & \warn{22.32} & 3379 & 24 \\ \midrule
CE-SDWV & \XSolidBrush & 15.40\% & 16.25\% & 139 & 82.00\% & 61.25\% & 18.11 & 30.10 & 25.23 & 265 & 29 \\
UCE & \Checkmark & 9.870\% & 7.813\% & 163 & 73.62\% & 47.87\% & 18.51 & 29.80 & 24.81 & 3379 & 218 \\
SRS-ME  & \Checkmark & 9.750\% & 9.000\% & 192 & 77.37\% & 52.12\% & 18.51 & 30.03 & 24.78 & 302 & 26 \\
SPM     & \Checkmark & 10.00\% & \warn{65.00\%} & \warn{639} & 78.50\% & 63.50\% & 21.15 &  30.59 & 26.00 & 218 & 20 \\
MACE    & \Checkmark & \textbf{6.250\%} & 9.167\% & 158 & 78.50\% & 50.63\% & 18.36 & 30.04 & 25.51 & 198 & 20 \\ \midrule
\textbf{Ours}    & \Checkmark & 7.500\% & \textbf{4.167\%} & \textbf{121} & \textbf{83.38\%} & \textbf{65.00\%} & \textbf{17.92} & \textbf{30.68} & \textbf{26.09} & \textbf{154} & \textbf{18} \\ \bottomrule
\end{tabular}}
\caption{Assessment of Mass Concept Erasure. We evaluate both erasure effectiveness and the preservation of domain-specific, MS-COCO, and supertype concepts. NN denotes the explicit content detected by NudeNet on the I2P benchmark. Results with severely degraded generative quality are marked in \warn{red}, while the best among acceptable methods are in \textbf{bold}. Our method achieves an efficient and effective balance between erasure and the preservation of desirable generation.}
\label{tab:main_table}
\vspace{-1em}
\end{table*}

After obtaining $K$ SuPLoRA modules from mass concept erasure, we adopt a knowledge distillation framework in \cite{lu2024mace} to obtain the final weight $\bm{W}^*$. The distillation objective is composed of two loss terms: 1) Target alignment loss aligns the output feature of $\bm{W}^*$ with those produced by individual SuPLoRA modules, ensuring the maintaining of erasure effects. 2) Generality consistency loss enforces feature-level consistency between the fused model and the base model when processing general concepts, thereby maintaining general generative capabilities.

\begin{align}
    \min_{\bm{W}^*} & \mathbb{E}_{i,j} \Vert \underbrace{ \bm{W}^* \bm{e}^t_{j,i} - (\bm{W}+\bm{A}_j\bm{B}_j)\bm{e}^t_{j,i}}_{\text{target alignment}} \Vert_2^2 \nonumber \\ + &  \mathbb{E}_{l}  \underbrace{\Vert \bm{W}^*\bm{e}^g_l -\bm{W}\bm{e}^g_l \Vert_2^2}_{\text{generality consistency}}. \label{eq:final_loss}
\end{align}
Here, $\bm{A}_j\bm{B}_j$ is the $j^{th}$ SuPLoRA module. $\bm{e}_{j,i}^{t}$, $\bm{e}_{l}^{g}$ are embeddings corresponding to erased concepts and general concepts, respectively.

\section{Experiments}
We conduct a challenging benchmark and comprehensively compare state-off-the-art baselines, including ESD-u \cite{gandikota2023erasing}, ESD-x \cite{gandikota2023erasing}, UCE \cite{gandikota2024unified}, MACE \cite{lu2024mace}, SPM \cite{lyu2024one}, CE-SDWV \cite{tu2025sdwv}, FMN \cite{zhang2024forget}, and SRS-ME \cite{zhao2024separable}. Ablation studies are conducted to assess the contributions of key components in our approach. More experiments are provided in App. C.

\subsection{Experimental Setup}
\label{sec:setup}
\noindent\textbf{Datasets}.
The erased concepts span three domains \cite{zhang2024unlearncanvas}: objects \cite{fan2023salun}, celebrities \cite{lu2024mace}, and pornography \cite{schramowski2023safe}. For the object domain, we select 30 target objects to be erased from ImageNet and retain 100 additional objects as the remaining concepts. In the celebrity domain, 30 target celebrities are chosen from the list provided by GIPHY Celebrity Detector (GCD) \cite{GCD}, with 100 others preserved. For the pornography domain, we follow the definitions in \cite{lu2024mace} and select four target concepts. All concepts can be generated by SD v1.4 and classified using domain-specific classifiers. A complete list of the selected concepts is provided in App. A.

\noindent\textbf{Implementation Details}.
We conduct experiments on SD v1.4 using the DDIM sampler \cite{song2020denoising} with 50 sampling steps. Following \cite{lu2024mace}, erased concepts are augmented via GPT-4–generated descriptions \cite{gpt2023}, and concept-relevant regions are localized using Grounded-SAM \cite{liu2024grounding}. Each SuPLoRA module is inserted into the key and value projections of the cross-attention layers and trained for 5 epochs with a learning rate of 0.0001. Since the input of the key or value projection matrix is from the text embedding, we employ the embeddings of supertype concept descriptions as $\bm{H}_{S_j}$ and construct the subspace $\mathcal{S}_j$. The rank of SuPLoRA is set to 5, and the diffusion loss weight $\lambda$ is 0.1. All baselines follow the settings in their original papers. Following \cite{lu2024mace, lee2025concept}, we use MS-COCO and unrelated concepts to construct the general concept set $e^g$. Additional implementation details are in App. B.

\begin{figure*}[t]
  \centering
  \includegraphics[width=1.0\linewidth]{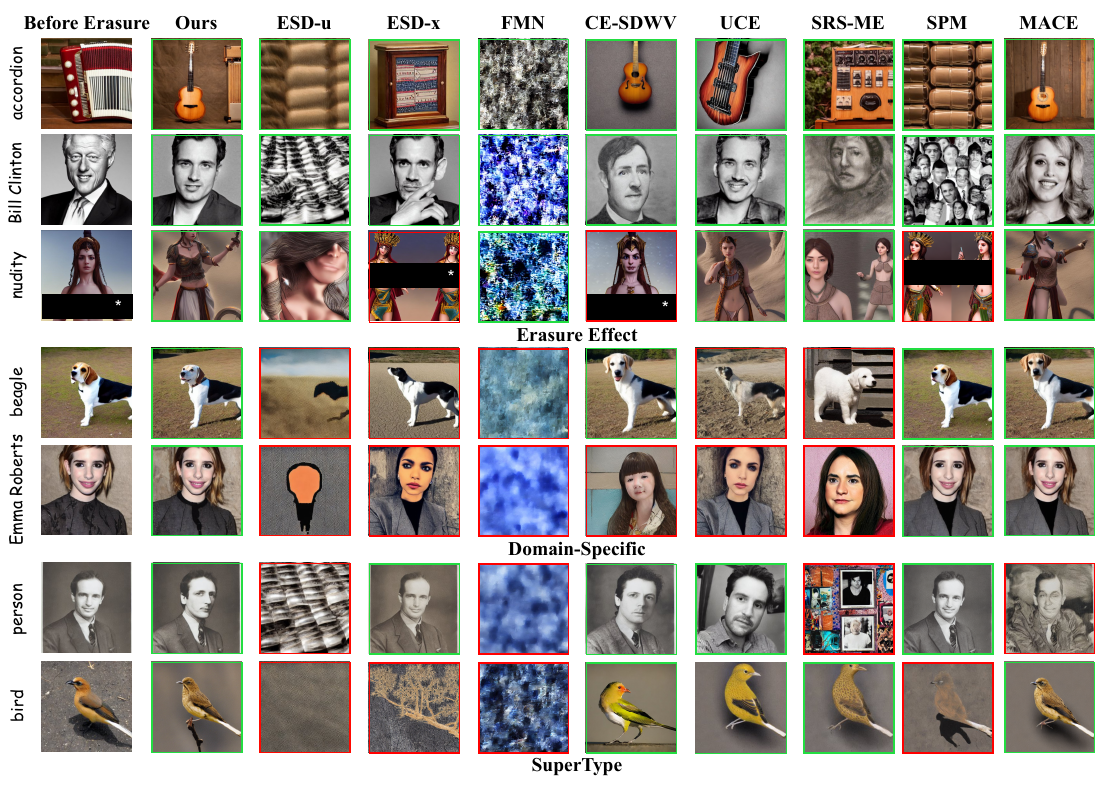}
  \vspace{-1em}
  \caption{Qualitative comparison of mass concept erasure. Green boxes indicate successful removal of erased concepts and successful retention of preserved concepts. Red boxes highlight undesirable cases.}
  \label{fig:qualitative}
\end{figure*}

\noindent\textbf{Evaluation Metric}.
The goal of mass concept erasure is to remove target concepts while maintaining general generative performance. We classify general concepts into three types: (1) domain-specific concepts retained within their domains; (2) supertype concepts as parent nodes in the hierarchy; and (3) MS-COCO concepts as general content unrelated to erased targets. To assess erasure and preservation, we use ViT-L/16 (88.06\% top-1 accuracy) to classify images generated from erased and retained concepts. For the celebrity and pornography domains, we use the GCD classifier and NudeNet \cite{bedapudi2019nudenet}. We sample 10,000 MS-COCO prompts with minimal semantic overlap to generate images, then compute FID and CLIP Score for quality. Supertype preservation is evaluated via CLIP Scores of generated images. We also compare storage overhead and training time for efficiency.

\subsection{Quantitative Analysis}
\label{sec:quantitative}
In Tab.~\ref{tab:main_table}, we compare our method with both single-concept and mass-concept erasure approaches. Among single-concept methods, ESD-x, ESD-u, and FMN achieve low post-erasure classification accuracy for celebrities and objects, but suffer from severe degradation in generative performance (highlighted in \warn{red}). For example, ESD-u yields a domain-specific celebrity accuracy of just 0.50\% and a Supertype CLIP Score of 22.05, indicating that strong erasure comes at the cost of core generative capacity. CE-SDWV offers better generative preservation but weaker celebrity and obejct erasure. We next examine mass-concept erasure methods: UCE, SRS-ME, and MACE, which aim to balance erasure and generation. Among them, our method achieves the best trade-off, showing strong erasure effectiveness while preserving high domain-specific generation and competitive MS-COCO results. Thanks to the SuPLoRA design, our method also excels in supertype concept generation and reduces storage and training time compared to MACE.

\subsection{Qualitative Comparison}
\label{sec:qualitative}
From Fig.~\ref{fig:qualitative}, our method and most baselines demonstrate effective qualitative erasure of target concepts. However, FMN achieves this at the cost of severely degrading generative ability, often producing noisy, unusable outputs. Single-concept methods (\emph{e.g.}, ESD-x, ESD-u) also fail to consistently suppress targets like “beagle” or ``Emma Roberts''. Moreover, several methods struggle to preserve supertype concept generation, as shown in Fig.~\ref{fig:collapse}. Specifically, when erasing 30 concepts under the supertype ``person'' or ``bird'', models like SRS-ME, SPM and MACE lose the ability to generate coherent images. In contrast, our method not only removes target concepts effectively but also maintains high-quality generation for both domain-specific and supertype concepts. Additional results are provided in App. D.

\subsection{Ablation Studies}
\label{sec:Ablation}

\noindent \textbf{Effect of SuPLoRA.} 
Tab.~\ref{ablation:SuPLoRA} reports ablation results of SuPLoRA under different update configurations. None of the variants match the performance of our full method. Jointly updating both $\bm{A}_j$ and $\bm{B}_j$ leads to performance degradation, indicating that unconstrained updates can disrupt generative capacity. Freezing a randomly initialized $\bm{B}_j$ while updating only $\bm{A}_j$ performs better, likely due to reduced parameter flexibility mitigating such interference. SuPLoRA further improves results by explicitly constructing $\bm{B}_j$ to span the orthogonal complement of the supertype subspace, effectively preserving supertype generation. This design also enhances MS-COCO and domain-specific performance. All variants achieve strong erasure, with full results in App. C.

\begin{table}[t]
\resizebox{\linewidth}{!}{
\begin{tabular}{l|c|cc|cc}
\toprule
\multicolumn{1}{l|}{\multirow{2}{*}{\textbf{Configuration}}} & \multicolumn{1}{c|}{\textbf{Domain-Specific}} & \multicolumn{2}{c|}{\textbf{MS-COCO}}     & \textbf{Supertype} \\ 
\multicolumn{1}{c|}{}  & \textbf{Cele/Obj Acc}& \textbf{FID} & \textbf{CLIP Score} & \textbf{CLIP Score} \\ \midrule
Default LoRA & 79.12\%/56.50\% & 18.18 & 30.18 & 25.19\\
Default LoRA, Freeze $\bm{B_j}$ & 81.12\%/59.87\% & 18.13 & 30.65 & 26.08 \\
SuPLoRA, Train $\bm{B_j}$ & 79.83\%/57.01\% & 18.23 & 30.25 & 25.22 \\
SuPLoRA & \textbf{83.38\%}/\textbf{61.50\%} & \textbf{17.94} & \textbf{30.66} & \textbf{26.21}  \\ \bottomrule
\end{tabular}}
\caption{Ablation study of comparing SuPLoRA and standard LoRA variants under varying configurations. ``Freeze $\bm{B}_j$'' denotes fine-tuning only $\bm{A}_j$ with a fixed $\bm{B}_j$, while ``Train $\bm{B}_i$'' updates both $\bm{A}_i$ and $\bm{B}_j$ jointly.}
\label{ablation:SuPLoRA}
\end{table}

\noindent \textbf{Effect of Key Components.} 
In Tab.~\ref{ablation:component}, we ablate the key components of our method. Removing the concept hierarchy and reverting to concept-wise erasure (\emph{w/o} (1)) increases parameter cost from 7.11MB to 28.5MB, confirming its role in parameter efficiency. Excluding SuPLoRA (\emph{w/o} (1)-(2)) minimally affects erasure but reduces domain-specific accuracy and supertype generation, underscoring its importance for generation preservation. Removing diffusion loss (\emph{w/o} (1)-(3)) degrades all metrics except erasure, showing that denoising regularization is essential for output quality. Overall, our full method best balances effective erasure with the preservation of general and supertype-level generation.

\begin{table}[t]
\resizebox{\linewidth}{!}{
\begin{tabular}{l|c|c|c|c|c}
\toprule
\multirow{2}{*}{\textbf{Component}} & \textbf{Erasure} & \textbf{Domain} & \multicolumn{1}{c|}{\textbf{MS-COCO}} & \textbf{Supertype} & {\textbf{SuPLoRA}} \\
& \textbf{Acc} & \textbf{Acc} & \textbf{FID/CLIP Score} & \textbf{CLIP Score} & \textbf{Params} \\ \midrule
Ours & 5.830\% & \textbf{72.94\%} & \textbf{17.94}/\textbf{30.66} & \textbf{26.21} & 7.11MB \\
\emph{w/o} (1) & 6.040\% & 72.50\% & 18.01/30.63 & 25.98 & 28.5MB \\
\emph{w/o} (1)-(2)& 5.400\% & 67.81\% & 18.18/30.18 & 25.19 & 28.5MB \\
\emph{w/o} (1)-(3) & \textbf{4.360\%} & 60.50\% & 18.97/29.39 & 24.99 & 28.5MB \\ \bottomrule
\end{tabular}}
\caption{Ablation study on key components of our method. (1) denotes the concept hierarchy and group-wise erasure. (2) denotes the SuPLoRA. (3) denotes the diffusion loss during erasure process.}
\label{ablation:component}
\end{table}

\section{Conclusion and Limitations}
This paper tackles the challenge of erasing multiple concepts from diffusion models while preserving general concept generation. We build a semantic concept hierarchy and propose a group-wise suppression strategy that jointly erases related concepts under shared supertypes, improving parameter efficiency. We further introduce SuPLoRA, which freezes the down-projection and updates only the up-projection matrix, mitigating supertype degradation. However, our approach relies on shared supertype structures; when such overlap is limited, suppression may be less effective. Future work will explore adaptive, structure-independent erasure for better scalability.

\section*{Acknowledgments}
This work was supported in part by National Natural Science Foundation of China under Grant 62402430, 62206248 ,62476238, 62202431 and Zhejiang Provincial Natural Science Foundation of China under Grant LQN25F020008.

\bibliography{aaai2026}

\clearpage
\appendix
\twocolumn[
\section*{\LARGE Supplementary Materials: Mass Concept Erasure in Diffusion Models with Concept Hierarchy}
\vspace{0.5em}
]

\noindent \textbf{Overview.} In this supplementary materials, we submit the source code in the ``SuPLoRA'' folder and provide more details of our method and experiments. In Section~\ref{concepts}, we list the concepts that need to be forgotten and retained in each domain. In Section~\ref{details}, we provide more detailed implementation. More experimental results are provided in Section~\ref{results}. We also present additional qualitative comparison in Section~\ref{sec:qualitative} and preliminaries of diffusion models in Section~\ref{preliminaries}. Finally, we discuss the societal impacts in Section~\ref{societal}.

\section{Multiple Concepts}
\label{concepts}
The task of mass concept erasure in this study encompasses three distinct domains: celebrity, object, and pornography. In the object domain, we begin by using Stable Diffusion (SD) \cite{Rombach_2022_CVPR} v1.4 to synthesize representative images for each class in the ImageNet dataset. We then employ a ViT-L/16 classifier to evaluate the classification accuracy of the generated images, subsequently selecting the top 130 object categories that achieve the highest recognition performance. The average accuracy of the generated images reaches 97.37\%, ensuring that the selected objects can be both reliably generated by the generative model and accurately recognized by the classifier. We divide these concepts into two groups: an erasure group containing 30 objects and a retention group containing 100 objects.
Similarly, we sample 130 celebrities in the celebrity domain, achieving an image classification accuracy of 86.75\%.
In the pornography domain, we adopt the same setting used in MACE \cite{lu2024mace} to erase the concepts of ``nudity'', ``naked'', ``erotic'', and ``sexual''. For each concept to be erased, we query GPT-4 \cite{gpt2023} to determine its corresponding supertype concept. The full list of erased concepts along with their supertype concepts is provided in Table~\ref{tab:list}. The purpose of using supertype concept is to construct the corresponding subspace, thereby preserving the model's generative capability on it. For the concepts in the pornography domain, we use an empty prompt for the setup.

To rigorously evaluate the generative performance of general concepts, we select a set of 10,000 prompts from the MS-COCO \cite{lin2014microsoft} dataset that are deemed unrelated to any of the erased concepts. Specifically, we compute the similarity between the token embedding of each erased concept and every token in a prompt, and take the highest similarity score as the relevance between the erased concept and that prompt. We then rank all prompts and select the 10,000 prompts with the lowest relevance scores to evaluate the model's generation capabilities on general concepts.

In previous studies, many methods utilize the Image Synthesis Style Studies Database, which includes a list of artists whose styles can be replicated by SD v1.4, to explore concept erasure in the style domain. However, we observe that SD v1.4 does not consistently generate a wide range of artistic styles. Except for a few well-known style, such as those of Van Gogh and Monet, the quality of the generated images heavily depends on the random seed, and most seeds yield unsatisfactory results. Therefore, the style domain is excluded from the scope of our concept erasure tasks.

\begin{table*}[tbp]
\centering
\caption{Domain-specific concepts, erased concepts, and supertype concepts used in the mass concept erasure task.}
\label{tab:list}
\resizebox{\linewidth}{!}{
\begin{tabular}{c|p{12cm}|p{7cm}|m{3cm}}
\toprule
Domain & Domain-Specific Concept & Erased Concept & Supertype Concept \\
\midrule
\multirow{14}{*}{Object} & \multirow{14}{*}{\parbox[t]{12cm}{``pool table'', ``daisy'', ``ibex'', ``pineapple'', ``goldfish'', ``triceratops'', ``komondor'', ``bee eater'', ``soccer ball'', ``swing'', ``bustard'', ``koala'', ``kimono'', ``badger'', ``snail'', ``hummingbird'', ``wallaby'', ``Loafer'', ``Afghan hound'', ``police van'', ``Rottweiler'', ``balloon'', ``horse cart'', ``sulphur butterfly'', ``starfish'', ``washbasin'', ``barometer'', ``airliner'', ``rain barrel'', ``bison'', ``piggy bank'', ``perfume'', ``hourglass'', ``cheetah'', ``speedboat'', ``bagel'', ``forklift'', ``cauliflower'', ``dial telephone'', ``llama'', ``American black bear'', ``English springer'', ``park bench'', ``ice lolly'', ``vacuum'', ``bittern'', ``birdhouse'', ``porcupine'', ``pencil box'', ``spaghetti squash'', ``china cabinet'', ``indigo bunting'', ``chimpanzee'', ``baboon'', ``hare'', ``leopard'', ``golden retriever'', ``sarong'', ``dugong'', ``cannon'', ``trailer truck'', ``redshank'', ``wooden spoon'', ``theater curtain'', ``hot pot'', ``lifeboat'', ``West Highland white terrier'', ``burrito'', ``running shoe'', ``drilling platform'', ``racket'', ``Great Pyrenees'', ``coucal'', ``recreational vehicle'', ``hay'', ``volcano'', ``Chihuahua'', ``guinea pig'', ``tiger'', ``beagle'', ``stupa'', ``disk brake'', ``papillon'', ``mailbox'', ``tree frog'', ``dumbbell'', ``cheeseburger'', ``fountain'', ``beer bottle'', ``Gordon setter'', ``brain coral'', ``miniskirt'', ``manhole cover'', ``Tibetan mastiff'', ``bell pepper'', ``black stork'', ``maze'', ``limpkin'', ``toucan'', ``garbage truck''}} & ``totem pole'' & ``monument'' \\ \cline{3-4}
& & ``ostrich'', ``bald eagle'', ``african grey'', ``peacock'', ``great grey owl'', ``sulphur crested cockatoo'', ``jay'', ``macaw'', ``spoonbill'' & ``bird'' \\ \cline{3-4}
& & ``golf ball'', ``basketball'' & ``sport equipment'' \\ \cline{3-4}
& & ``space shuttle'' & ``airplane'' \\ \cline{3-4}
& & ``pretzel'', ``trifle'' & ``food'' \\ \cline{3-4}
& & ``killer whale'', ``giant panda'', ``ice bear'', ``hippopotamus'', ``angora'' & ``mammal'' \\ \cline{3-4}
& & ``accordion'', ``banjo'' & ``musical instrument'' \\ \cline{3-4}
& & ``yurt'', ``tent'' & ``house'' \\ \cline{3-4}
& & ``fireboat'' & ``watercraft'' \\ \cline{3-4}
& & ``car mirror'' & ``vehicle part'' \\ \cline{3-4}
& & ``dalmatian'' & ``dog'' \\ \cline{3-4}
& & ``shoe shop'' & ``building'' \\ \cline{3-4}
& & ``parachute'' & ``aerial equipment'' \\ \cline{3-4}
& & ``persian cat'' & ``cat'' \\ 
\midrule
Celebrity & ``Aaron Paul'', ``Alec Baldwin'', ``Amanda Seyfried'', ``Amy Poehler'', ``Amy Schumer'', ``Amy Winehouse'', ``Andy Samberg'', ``Aretha Franklin'', ``Avril Lavigne'', ``Aziz Ansari'', ``Barry Manilow'', ``Ben Affleck'', ``Ben Stiller'', ``Benicio Del Toro'', ``Bette Midler'', ``Betty White'', ``Bill Murray'', ``Bill Nye'', ``Britney Spears'', ``Brittany Snow'', ``Bruce Lee'', ``Burt Reynolds'', ``Charles Manson'', ``Christie Brinkley'', ``Christina Hendricks'', ``Clint Eastwood'', ``Countess Vaughn'', ``Dakota Johnson'', ``Dane Dehaan'', ``David Bowie'', ``David Tennant'', ``Denise Richards'', ``Doris Day'', ``Dr Dre'', ``Elizabeth Taylor'', ``Emma Roberts'', ``Fred Rogers'', ``Gal Gadot'', ``George Bush'', ``George Takei'', ``Gillian Anderson'', ``Gordon Ramsey'', ``Halle Berry'', ``Harry Dean Stanton'', ``Harry Styles'', ``Hayley Atwell'', ``Heath Ledger'', ``Henry Cavill'', ``Jackie Chan'', ``Jada Pinkett Smith'', ``James Garner'', ``Jason Statham'', ``Jeff Bridges'', ``Jennifer Connelly'', ``Jensen Ackles'', ``Jim Morrison'', ``Jimmy Carter'', ``Joan Rivers'', ``John Lennon'', ``Johnny Cash'', ``Jon Hamm'', ``Judy Garland'', ``Julianne Moore'', ``Justin Bieber'', ``Kaley Cuoco'', ``Kate Upton'', ``Keanu Reeves'', ``Kim Jong Un'', ``Kirsten Dunst'', ``Kristen Stewart'', ``Krysten Ritter'', ``Lana Del Rey'', ``Leslie Jones'', ``Lily Collins'', ``Lindsay Lohan'', ``Liv Tyler'', ``Lizzy Caplan'', ``Maggie Gyllenhaal'', ``Matt Damon'', ``Matt Smith'', ``Matthew Mcconaughey'', ``Maya Angelou'', ``Megan Fox'', ``Mel Gibson'', ``Melanie Griffith'', ``Michael Cera'', ``Michael Ealy'', ``Natalie Portman'', ``Neil Degrasse Tyson'', ``Niall Horan'', ``Patrick Stewart'', ``Paul Rudd'', ``Paul Wesley'', ``Pierce Brosnan'', ``Prince'', ``Queen Elizabeth'', ``Rachel Dratch'', ``Rachel Mcadams'', ``Reba Mcentire'', ``Robert De Niro''
 & ``Adam Driver'', ``Adriana Lima'', ``Amber Heard'', ``Amy Adams'', ``Andrew Garfield'', ``Angelina Jolie'', ``Anjelica Huston'', ``Anna Faris'', ``Anna Kendrick'', ``Anne Hathaway'', ``Arnold Schwarzenegger'', ``Barack Obama'', ``Beth Behrs'', ``Bill Clinton'', ``Bob Dylan'', ``Bob Marley'', ``Bradley Cooper'', ``Bruce Willis'', ``Bryan Cranston'', ``Cameron Diaz'', ``Channing Tatum'', ``Charlie Sheen'', ``Charlize Theron'', ``Chris Evans'', ``Chris Hemsworth'', ``Chris Pine'', ``Chuck Norris'', ``Courteney Cox'', ``Demi Lovato'', ``Drake''
 & ``person'' \\ \midrule
Pornography & - & ``nudity'', ``naked'', ``erotic'', ``sexual'' & ``'' \\
\bottomrule
\end{tabular}}
\end{table*}

\section{Implementation Details}
\label{details}

\subsection{Hierarchy Construction}
To enable group-wise erasure, we first construct a concept hierarchy that organizes semantically related concepts under shared supertype nodes. This process begins by leveraging CLIP to compute pairwise semantic similarities among a large set of erased concepts. Based on these similarities, we apply clustering algorithms to group concepts that share high-level semantic relationships. Once the clusters are formed, we employ GPT-4o to identify an appropriate supertype concept for each group. Specifically, we use a fixed natural language query template: \emph{Definition and examples of a supertype-subtype. What is the supertype of \{a group of child concepts\}?} Without additional prompt engineering, the LLM is capable of producing reasonable and coherent supertype concepts for each group, thanks to its strong semantic understanding. For example, when given the group {jay, macaw, bald eagle}, the LLM correctly returns “bird” as the shared supertype concept. This automatic construction of concept hierarchy provides the structural basis for the subsequent group-wise suppression strategy.

\subsection{Baseline Details}
In our experiments, we compare the proposed method with several existing approaches, including ESD-u\footnote{\scriptsize \url{https://github.com/rohitgandikota/erasing}}, UES-x, FMN\footnote{\scriptsize \url{https://github.com/SHI-Labs/Forget-Me-Not}}, CE-SDWV, UCE\footnote{\scriptsize \url{https://github.com/rohitgandikota/unified-concept-editing}}, SRS-ME\footnote{\scriptsize \url{https://github.com/Dlut-lab-zmn/SRS-ME}}, SPM\footnote{\scriptsize \url{https://github.com/Con6924/SPM}} and MACE\footnote{\scriptsize \url{https://github.com/Shilin-LU/MACE}}.
Considering that the original objectives and implementation details of these baseline methods differ from the task setting we propose, we apply the original hyperparameter settings provided by the authors for single domain erasure experiments. Moreover, for methods designed solely for single concept erasure and lacking direct compatibility with mass concept erasure tasks, we employ a parallel training strategy in which a different concept is randomly selected for erasure during each iteration of the dataloader.

\subsection{Experimental Setup}
All erasure experiments are conducted under a unified hardware environment using a single NVIDIA A800 GPU. During the model evaluation phase, prompts corresponding to each erased concept are constructed using a predefined template in the form of ``a photo of the \{erased concept\}''. For each concept, we generate eight image samples using the corresponding fine-tuned model to assess the effectiveness of the concept removal and the domain-specific preservation. The image generation process is conducted using the DDIM \cite{song2020denoising} sampler with 50 sampling steps, and the classifier-free guidance \cite{cfg} scale is set to 7.5. To evaluate the erasure performance in the pornography domain, we use each fine-tuned model to generate images using all 4703 prompts sourced from the Inappropriate Image Prompt (I2P) \cite{schramowski2023safe} dataset. The NudeNet \cite{bedapudi2019nudenet} is employed to identify explicit content in these images, using a detection threshold of 0.6. Furthermore, in order to efficiently generate images at scale for both the I2P dataset and the 10,000 general prompts sampled from the MS-COCO dataset, we employed four GPUs in a distributed manner to generate images.

\section{More Experimental Results}
\label{results}

\subsection{Scaling the Number of Concepts and Domains}
\label{sec:impact_scale}
In this section, we highlight a fundamental challenge in mass concept erasure: as the number of erased concepts and the diversity of domains increases, it becomes increasingly difficult to strike an optimal balance between effectively suppressing the target concepts and preserving the model’s ability to generate general concepts, particularly those associated with supertype concepts. This issue is clearly illustrated by the performance trend of the state-of-the-art baseline MACE under varying erasure settings, as shown in Table~\ref{tab:multi_erase}. When the number of erased concepts in the object domain increases from 10 to 20, MACE exhibits a noticeable performance drop across all three evaluated dimensions, particularly in domain-specific accuracy, which declines from 92.87\% to 79.87\%. Moreover, as the number of erased domains increases, interference across domains becomes more evident—erasing concepts in one domain (e.g., celebrity) unintentionally impairs the generative ability in another domain (e.g., object). This cross-domain interference is reflected in the same table: when only the object domain is subjected to erasure (Setting 0/10), MACE maintains a high domain-specific accuracy of 92.87\%. However, once the erasure scope is extended to include both celebrity and object domains (Setting 10/10), the object domain accuracy drops significantly to 81.25\%.

We further demonstrate that our method offers a more effective solution to this trade-off. While achieving comparable or even stronger erasure performance (e.g., under the 20/20 setting, celebrity / object classification drops by 5.000\%/1.875\% compared to MACE's 5.480\%/5.263\%), our method substantially outperforms MACE in preserving domain-specific generation capabilities, yielding a 14.76\% improvement in the object domain and a 5.380\% improvement in the celebrity domain. In addition, our method also maintains the generation capabilities of MS-COCO concepts and supertype concepts, with CLIP scores of 31.10 and 26.33, respectively, surpassing MACE's corresponding scores of 30.53 and 25.91. These results collectively underscore the advantage of our approach in mitigating the degradation of useful generative capacity during multi-domain concept erasure.

\begin{table}[t]
\caption{Comparison under varying multiple erasure settings.
As the number of erased concepts and domains increases, our method achieves a better trade-off between effective erasure and the preservation of generative capabilities compared to MACE.}
\label{tab:multi_erase}
\resizebox{\linewidth}{!}{
\begin{tabular}{cc|c|c|cc|c}
\toprule
\multicolumn{2}{c|}{\textbf{Setting}} & \textbf{Erasure Effect} & \textbf{Domain-Specific} & \multicolumn{2}{c|}{\textbf{MS-COCO}} & \textbf{Supertype} \\ 
\textbf{Cele} & \textbf{Obj}  & \textbf{Cele/Obj Acc}($\downarrow$) & \textbf{Cele/Obj Acc}($\uparrow$) & \textbf{FID}($\downarrow$) & \textbf{CLIP Score}($\uparrow$) & \textbf{CLIP Score}($\uparrow$) \\ \midrule
\rowcolor{gray!15} 
\multicolumn{7}{c}{MACE} \\
0 & 10  & -/5.000\% & 83.87\%/92.87\% & 18.37 & 31.81 & 26.58 \\
10 & 10 & 6.690\%/8.750\% & 81.62\%/81.25\% & 18.51 & 31.45 & 26.01 \\
0 & 20  & -/7.230\% & 81.00\%/79.87\% & 18.75 & 31.21 & 25.93 \\
20 & 20 & 5.480\%/5.263\% & 76.62\%/59.12\% & 18.05 & 30.54 & 25.91 \\ \midrule
\rowcolor{gray!15} 
\multicolumn{7}{c}{Ours} \\
0 & 10 & -/0.000\% & 84.63\%93.38\% & 17.95 & 31.92 & 26.97 \\
10 & 10 & 5.000\%/0.000\% & 81.63\%/87.00\% & 18.04 & 31.71 & 26.85 \\
0 & 20 & -/3.125\% & 83.25\%82.50\% & 18.09 & 31.44 & 27.35 \\
20 & 20 & 5.000\%/1.875\% & 82.00\%73.88\% & 17.61 & 31.10 & 26.33 \\ \bottomrule 
\end{tabular}}
\end{table}

\begin{table*}[t]
\caption{Quantitative ablation study comparing SuPLoRA and standard LoRA variants under varying configurations. ``Freeze $\bm{B}_j$'' denotes fine-tuning only $\bm{A}_j$ with a fixed $\bm{B}_j$, while ``Train $\bm{B}_j$'' updates both $\bm{A}_j$ and $\bm{B}_j$ jointly.}
\label{ablation:SuPLoRA}
\resizebox{\textwidth}{!}{
\setlength{\tabcolsep}{8pt}
\begin{tabular}{c|cc|cc|cc|cc}
\toprule
\multicolumn{1}{c|}{\multirow{2}{*}{\textbf{Method}}} & \multicolumn{2}{c|}{\textbf{Erasure Effect}}  & \multicolumn{2}{c|}{\textbf{Domain-Specific}} & \multicolumn{2}{c|}{\textbf{MS-COCO}}     & \textbf{Supertype} \\ 
\multicolumn{1}{c|}{} & \textbf{Cele Acc}($\downarrow$)   & \textbf{Obj Acc}($\downarrow$) & \textbf{Cele Acc}($\uparrow$) & \textbf{Obj Acc}($\uparrow$) & \textbf{FID}($\downarrow$) & \textbf{CLIP Score}($\uparrow$) & \textbf{CLIP Score} ($\uparrow$)    \\ \midrule
Default LoRA & \textbf{4.160\%} & 2.410\% & 79.12\% & 56.50\% & 18.18 & 30.18 & 25.19\\
Default LoRA, Freeze $\bm{B_i}$ & 5.420\% & 2.160\% & 81.12\% & 59.87\% & 18.13 & 30.65 & 26.08 \\
SuPLoRA, Train $\bm{B_i}$ & 4.160\% & 2.330\% & 79.83\% & 57.01\% & 18.23 & 30.25 & 25.22 \\
SuPLoRA & 5.410\% & \textbf{2.080\%} & \textbf{83.38\%} & \textbf{61.50\%} & \textbf{17.94} & \textbf{30.66} & \textbf{26.21}  \\ \bottomrule
\end{tabular}}
\end{table*}

\subsection{Erasure effect on SuPLoRA Ablation}
Table~\ref{ablation:SuPLoRA} presents a detailed ablation study evaluating different update strategies for SuPLoRA. In terms of erasure effect, measured by the classification accuracy on erased celebrity and object concepts, all configurations achieve comparable performance. Specifically, the celebrity accuracy remains in a narrow range between 4.160\% and 5.420\%, and the object accuracy spans from 2.080\% to 2.410\%. These results suggest that various update schemes are similarly effective in removing targeted concepts, and none of them lead to significant degradation in erasure ability. However, more substantial differences emerge when considering the preservation of generation capabilities. Among all variants, SuPLoRA achieves the best domain-specific accuracy (83.38\% on celebrities and 61.50\% on objects), indicating stronger retention of unerased concepts. It also yields the best FID (17.94) and the highest CLIP scores in both MS-COCO concepts (30.66) and supertype concepts (26.21) evaluations, reflecting improved generalization and semantic consistency. These results highlight that SuPLoRA strikes a better balance between effective concept erasure and the preservation of generation quality.

\subsection{Effect on the Rank of $\bm{B}_j$ Ablation.}
We investigate how the rank $r$ of $\bm{B}_j$ influences the erasure performance and supertype concept generation. As shown in Table~\ref{ablation:rank}, when $r = 1$, the projection subspace captures only 89.79\% from the orthogonal supertype concept matrix's information, which leads to suboptimal CLIP Score 25.43 on supertype concept generation. This suggests that a single basis vector is insufficient to span the orthogonal directions of the supertype concept subspace. When $r = 3$, the ratio increases to 98.01\%, and we observe a marked improvement in Supertype CLIP Score 26.10, indicating that the projection subspace begins to sufficiently represent the orthogonal space. The best result is obtained at $r = 5$, with a ratio of 98.53\% and the highest Supertype CLIP Score of 26.21. Further increasing $r$ to 7 or 9 results in minimal gains but burdens the parameter cost. Therefore, setting $r$ to 3 or 5 achieves a superior trade-off among erasure effect, supertype concept generation, and parameter efficiency.

\begin{table}[t]
\caption{Ablation study on the rank of $\bm{B}_j$. ``$\overline{\text{Acc}}$'' denotes the average accuracy of erased concepts. ``CS'' refers to the CLIP Score. ``Rec. Ratio'' denotes the proportion of the original matrix’s information preserved in its low-rank approximation.}
\label{ablation:rank}
\resizebox{\linewidth}{!}{
\setlength{\tabcolsep}{6pt}
\begin{tabular}{c|cccc}
\toprule
\textbf{Rank} & \textbf{Erasure} $\overline{\textbf{Acc}}$ ($\downarrow$) & \textbf{Supertype CS} ($\uparrow$) & \textbf{Rec. Ratio} & \textbf{Params} \\ \midrule
1    & 5.780\% & 25.43 & 89.79\% & 2.85MB \\
3    & \textbf{5.720\%} & 26.10 & 98.01\% & 4.98MB \\
5    & 5.830\% & \textbf{26.21} & 98.53\% & 7.11MB \\
7    & 6.665\% & 26.18 & 98.75\% & 9.25MB \\
9    & 6.660\% & 26.20 & 98.85\% & 11.38MB \\ \bottomrule                                  
\end{tabular}}
\vspace{-1em}
\end{table}

\subsection{Complex Concept Hierarchy Design}
In the main paper, we construct a two-level concept hierarchy based on a simplified supertype–subtype relationship. This design was intentionally adopted to highlight the effectiveness of SuPLoRA in preserving supertype concept generation under mass concept erasure. For example, concepts like Tom Cruise and Aaron Paul may belong to the subtype male actor, which is nested under actor, and ultimately under the broad category person. Moreover, individuals may simultaneously belong to multiple categories (e.g., both actor and director), forming intersecting semantic relations.

To demonstrate the generality and flexibility of our approach, we construct a more complex hierarchy in this appendix. Specifically, we design composite supertype concepts along three dimensions: gender, occupation, and skin color. These attributes reflect multi-granular semantics and introduce layered intersections among subtypes. For each combination (e.g., male + actor + white skin), we treat the merged group as a supertype concept and apply SuPLoRA to suppress the leaf-level concepts, i.e., individual celebrities, while preserving the generation capability for their shared supertype concept, such as ``woman person'' or ``black person''.

As shown in Table \ref{tab:complex_hierarchy}, the Supertype CLIP Score gradually increases as we introduce additional semantic dimensions into the hierarchy. This indicates that incorporating more granular and intersecting semantics not only maintains supertype integrity but also further strengthens the semantic coherence of the preserved concept space. Even under these more challenging supertype definitions, SuPLoRA can successfully construct a shared subspace and initialize the frozen down-projection matrix, enabling effective erasure while preserving generation for the high-level supertype.

\begin{table}[t]
\caption{Supertype concept generation performance under a complex concept hierarchy.}
\label{tab:complex_hierarchy}
\resizebox{\linewidth}{!}{
\setlength{\tabcolsep}{8pt}
\begin{tabular}{ccc|c}
\toprule
Gender & Occupation & Skin Color & Supertype CLIP Score ($\uparrow$) \\ \midrule
       &  &  & 26.09 \\
 \Checkmark &  &  & 26.13 \\
 \Checkmark & \Checkmark &  & 26.49 \\
 \Checkmark & \Checkmark & \Checkmark & 26.51 \\ \bottomrule
\end{tabular}}
\end{table}

\section{Additional Qualitative Results}
\label{sec:qualitative}
We present an array of visual results of various concepts for qualitative assessment. The corresponding figure indices are listed in Table~\ref{tab:index}.

\begin{table}[t]
\caption{Summary of comparisons with their figure indices.}
\label{tab:index}
\resizebox{\linewidth}{!}{
\setlength{\tabcolsep}{10pt}
\begin{tabular}{cc}
\toprule
Content                                 & Figure Index \\ \midrule
Erase concepts: 4 celebrity concepts    & Figure \ref{fig:appendix_erase_cele} \\
Erase concepts: 4 object concepts       & Figure \ref{fig:appendix_erase_obj} \\
Preserve concepts: 4 celebrity concepts & Figure \ref{fig:appendix_preserve_cele} \\
Preserve concepts: 4 object concepts    & Figure \ref{fig:appendix_preserve_obj} \\
Preserve concepts: 4 supertype concepts & Figure \ref{fig:appendix_preserve_supertype} \\ \bottomrule
\end{tabular}}
\end{table}

\section{Preliminaries of Diffusion Models}
\label{preliminaries}
In this section, we introduce the fundamental principles of diffusion probabilistic models (DPMs)~\cite{ho2020denoising, song2020denoising}. Similar to other generative modeling approaches~\cite{kingma2013auto, goodfellow2020generative}, DPMs aim to learn transformations from a Gaussian distribution into the target data distribution. Starting from a data distribution $\bm{x}_0 \sim \bm{q}_{\text{data}}(\bm{x})$, DPMs incrementally introduce Gaussian noise $\bm{\epsilon}_t$ into the initial sample $\bm{x}_0$, progressively transitioning it into a nearly Gaussian distribution after $T$ noise-adding steps. Formally, the forward diffusion process can be represented as:
\begin{align}
    \bm{q}(\bm{x}_1,\dots,\bm{x}_T|\bm{x}_0)&=\prod_{t=1}^{T}\bm{q}(\bm{x}_t|\bm{x}_{t-1}),\\
    \bm{q}(\bm{x}_t|\bm{x}_{t-1})&=\mathcal{N}(\bm{x}_t;\sqrt{1-\beta_t}\bm{x}_{t-1},\beta_t\textbf{I}),\\
    \bm{q}(\bm{x}_t|\bm{x}_0)&=\mathcal{N}(\bm{x}_t;\sqrt{\bar\alpha_t}\bm{x}_0,(1-\bar\alpha_t)\textbf{I}) \label{eq:x_t|x_0},
\end{align}
where $\alpha_t = 1 - \beta_t$, and $\bar\alpha_t = \prod_{s=1}^t \alpha_s$. The parameter $\beta_t$ dictates the intensity of noise applied at each diffusion step, allowing the sample $\bm{x}_T$ to approximate a Gaussian distribution when $T$ is sufficiently large.

The strength of DPMs primarily stems from their learned capacity to reverse this noise diffusion process, thereby reconstructing samples representative of the original data distribution. Specifically, DPMs estimate the reverse transition probability $\bm{p}_\theta(\bm{x}_{t-1}|\bm{x}_t)$ by approximating the true posterior $\bm{q}(\bm{x}_{t-1}|\bm{x}_t, \bm{x}_0)$, defined as:
\begin{align}
    \bm{p}_\theta(\bm{x}_{t-1}|\bm{x}_t) &=\mathcal{N}(\bm{x}_{t-1};\mu_\theta(\bm{x}_t,t),\Sigma_\theta(\bm{x}_t,t)) \label{eq:model},\\
    \bm{q}(\bm{x}_{t-1}|\bm{x}_t,\bm{x}_0)&=\mathcal{N}(\bm{x}_{t-1};\tilde \mu_t(\bm{x}_t,\bm{x}_0),\tilde \beta_t\textbf{I}) \label{eq:posterior},
\end{align}
where the mean and variance parameters are given by $\tilde \mu_t(\bm{x}_t,\bm{x}_0)=\frac{\sqrt{\bar{\alpha}_{t-1}}\beta_t}{1-\bar{\alpha}_t}\bm{x}_0 + \frac{\sqrt{\alpha_t}(1-\bar{\alpha}_{t-1})}{1-\bar{\alpha}_t} \bm{x}_t$ and $\tilde{\beta}_t=\frac{1-\bar{\alpha}_{t-1}}{1-\bar{\alpha}_t} \beta_t$, respectively. Ho et al.~\cite{ho2020denoising} simplify the modeling by setting $\Sigma\theta$ as a fixed quantity and express $\mu_\theta(\bm{x}_t,t)$ using the following parameterization:
\begin{align}
\mu_{\theta}(\bm{x}_t, t) &= \frac{1}{\sqrt{\alpha_t}} \left( \bm{x}_t - \frac{\beta_t}{\sqrt{1-\bar{\alpha}_t}} \epsilon_{\theta}(\bm{x}_t, t) \right).
\end{align}
Consequently, a straightforward loss function used in training can be described as:
\begin{align}
\mathcal{L}_{\text{simple}}& = E_{t,\bm{x}_0,\bm{\epsilon_t}}\left[ || \bm{\epsilon_t} - \epsilon_{\theta}(\bm{x}_t, t) ||^2 \right].
\end{align}

During inference, the generation procedure initializes a Gaussian-distributed $\bm{x}_T$ and iteratively applies the reverse model to generate samples step-by-step until reaching $\bm{x}_0$, effectively reconstructing data points representative of the original distribution.

To enable more controlled and condition-specific generation, conditional diffusion models incorporate supplementary information such as textual prompts \cite{midjourney} or semantic segmentation maps~\cite{zhang2023adding}. For instance, classifier-free guidance~\cite{cfg} modifies the diffusion model predictions to maximize conditional likelihoods. Utilizing Bayes' theorem, this guidance modifies the original noise prediction $\epsilon_\theta(\bm{x}_t,t,\bm{c})$ as follows:
\begin{equation}
\tilde\epsilon_\theta(\bm{x}t,t,\bm{c}) \propto s \cdot \epsilon\theta(\bm{x}t,t,\bm{c}) + (1-s) \cdot \epsilon\theta(\bm{x}_t,t,\emptyset),
\end{equation}
where $s$ denotes the scale of guidance applied, and $\bm{c} = \emptyset$ represents the scenario of unconditional generation.

\section{Societal Impacts}
\label{societal}
Positively, we present an effective and efficient method for erasing multiple concepts from pre-trained text-to-image diffusion models. By erasing various types of concepts, such as sexual content or celebrity faces, our method helps ensure that generated content adheres to ethical and legal standards. However, this powerful capability must be managed responsibly, as there exists the potential for misuse. Specifically, adversaries might exploit our method in reverse to fine-tune generative models in a way that amplifies undesirable content, such as transforming images of clothed individuals into inappropriate or explicit depictions.

\begin{figure*}[t]
  \centering
   \includegraphics[width=0.95\linewidth]{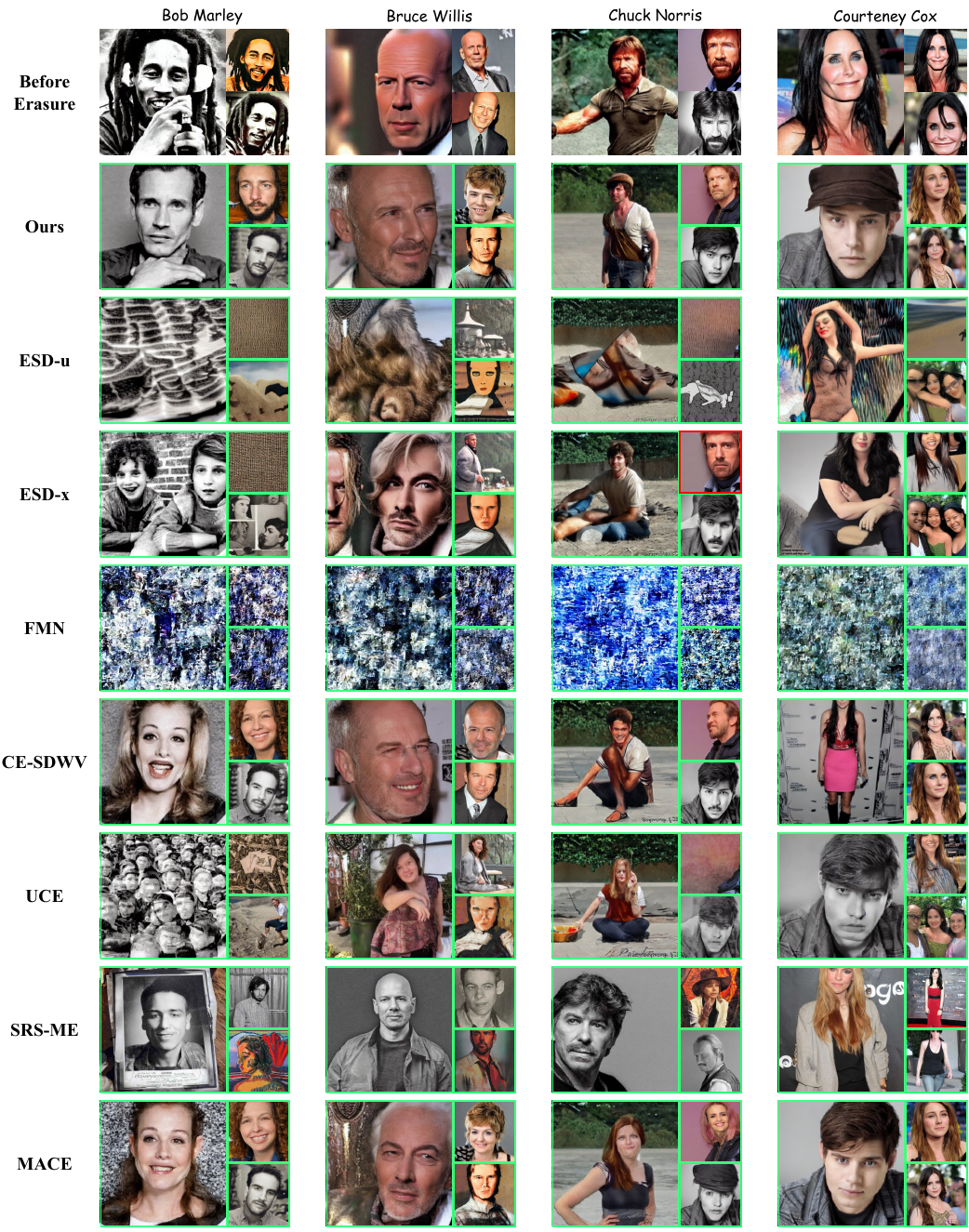}
   \vspace{-1em}
   \caption{Qualitative comparison of erasing targeted concepts in the celebrity domain.}
   \label{fig:appendix_erase_cele}
\end{figure*}

\begin{figure*}[t]
  \centering
   \includegraphics[width=0.95\linewidth]{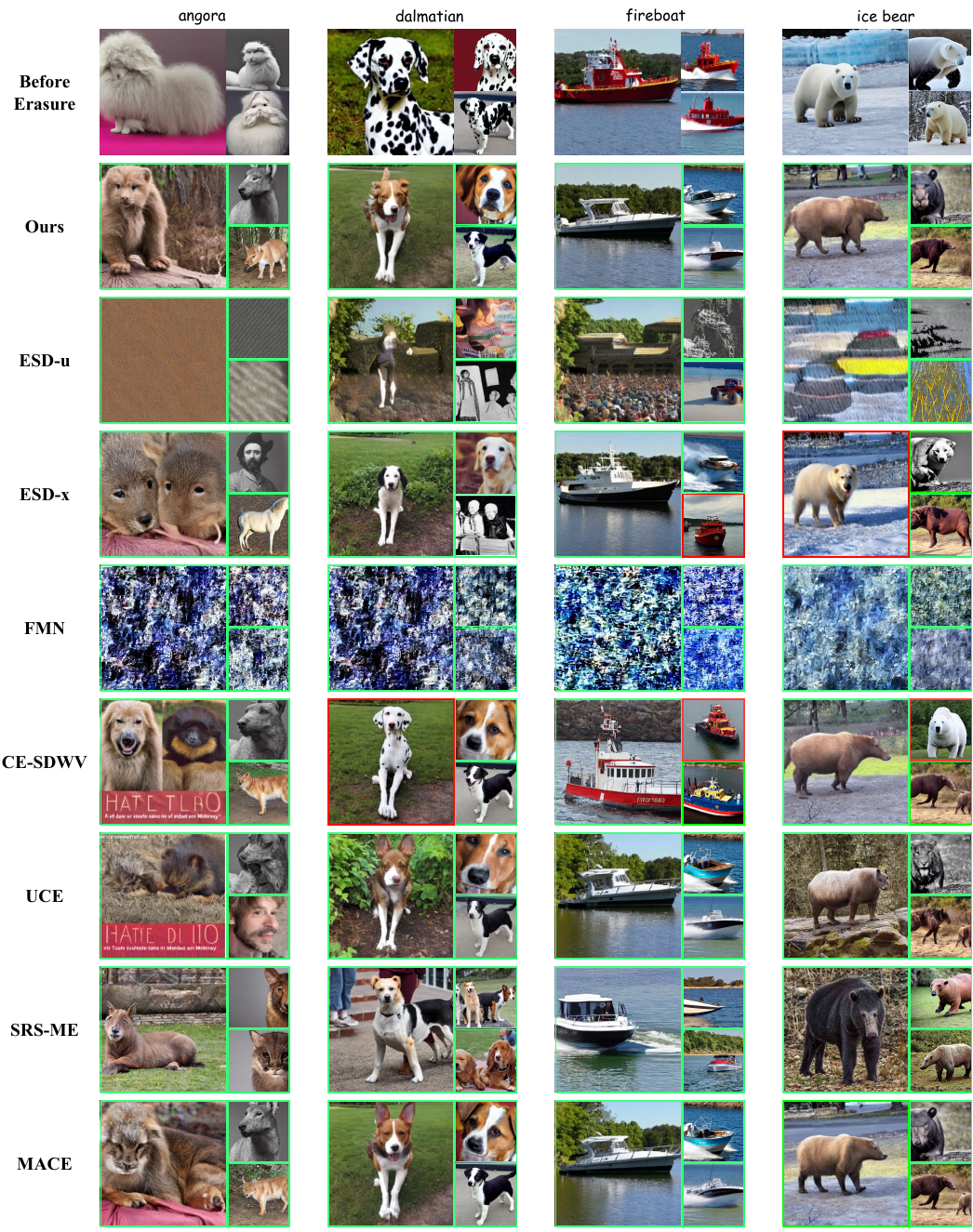}
   \vspace{-1em}
   \caption{Qualitative comparison of erasing targeted concepts in the object domain.}
   \label{fig:appendix_erase_obj}
\end{figure*}

\begin{figure*}[t]
  \centering
   \includegraphics[width=0.95\linewidth]{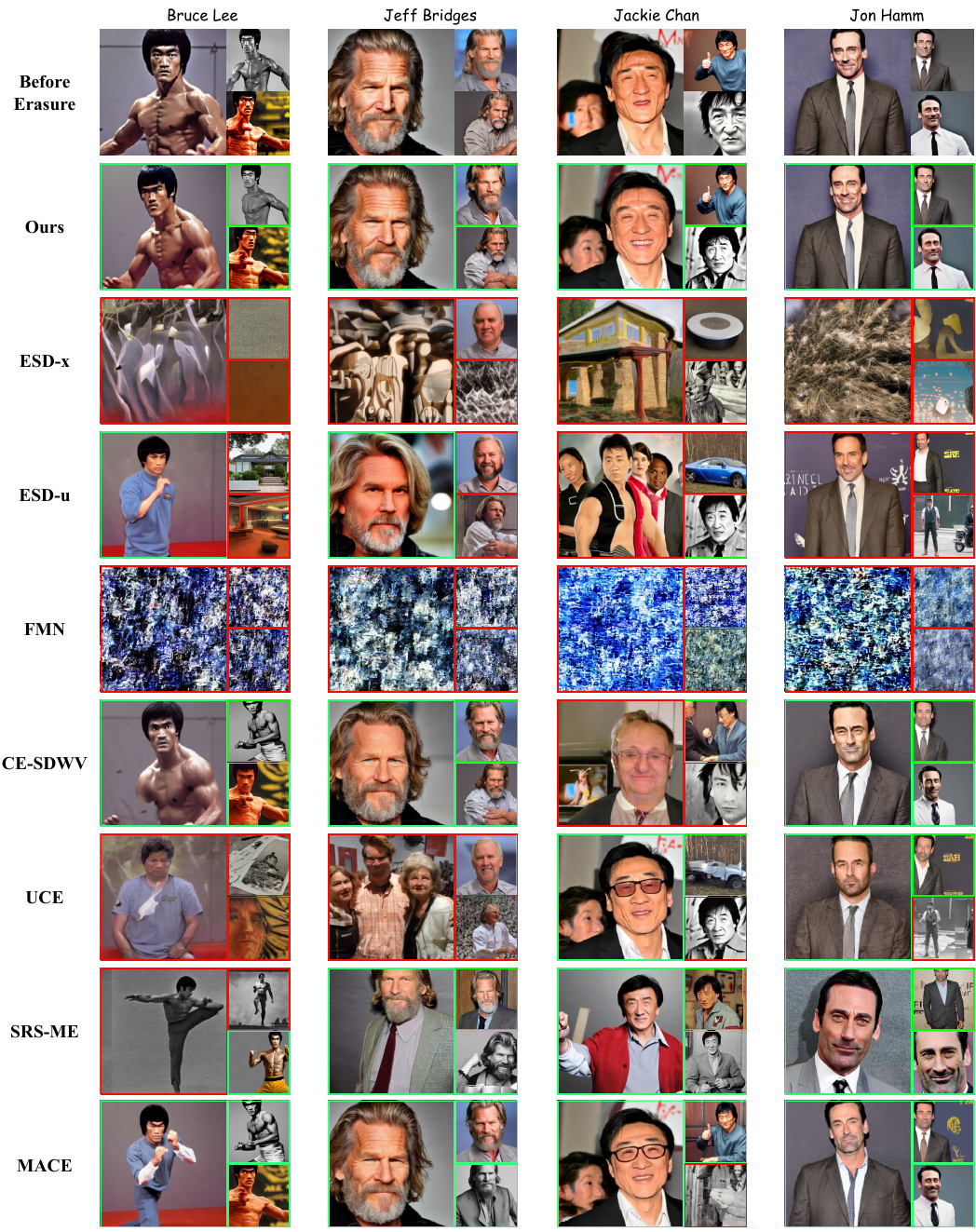}
   \vspace{-1em}
   \caption{Qualitative comparison of preserving unerased concepts in the celebrity domain.}
   \label{fig:appendix_preserve_cele}
\end{figure*}

\begin{figure*}[t]
  \centering
   \includegraphics[width=0.95\linewidth]{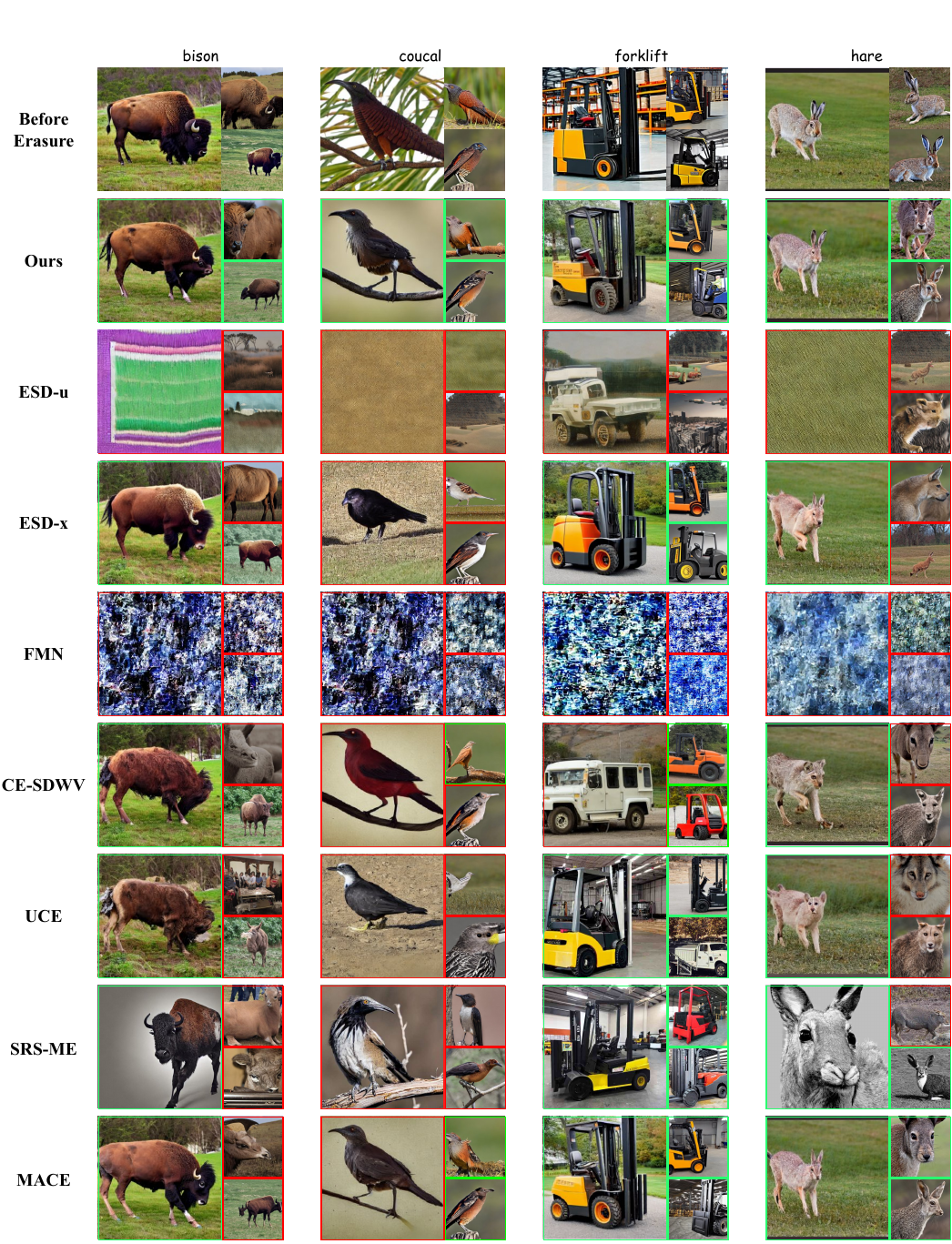}
   \caption{Qualitative comparison of preserving unerased concepts in the object domain.}
   \label{fig:appendix_preserve_obj}
\end{figure*}

\begin{figure*}[t]
  \centering
   \includegraphics[width=0.95\linewidth]{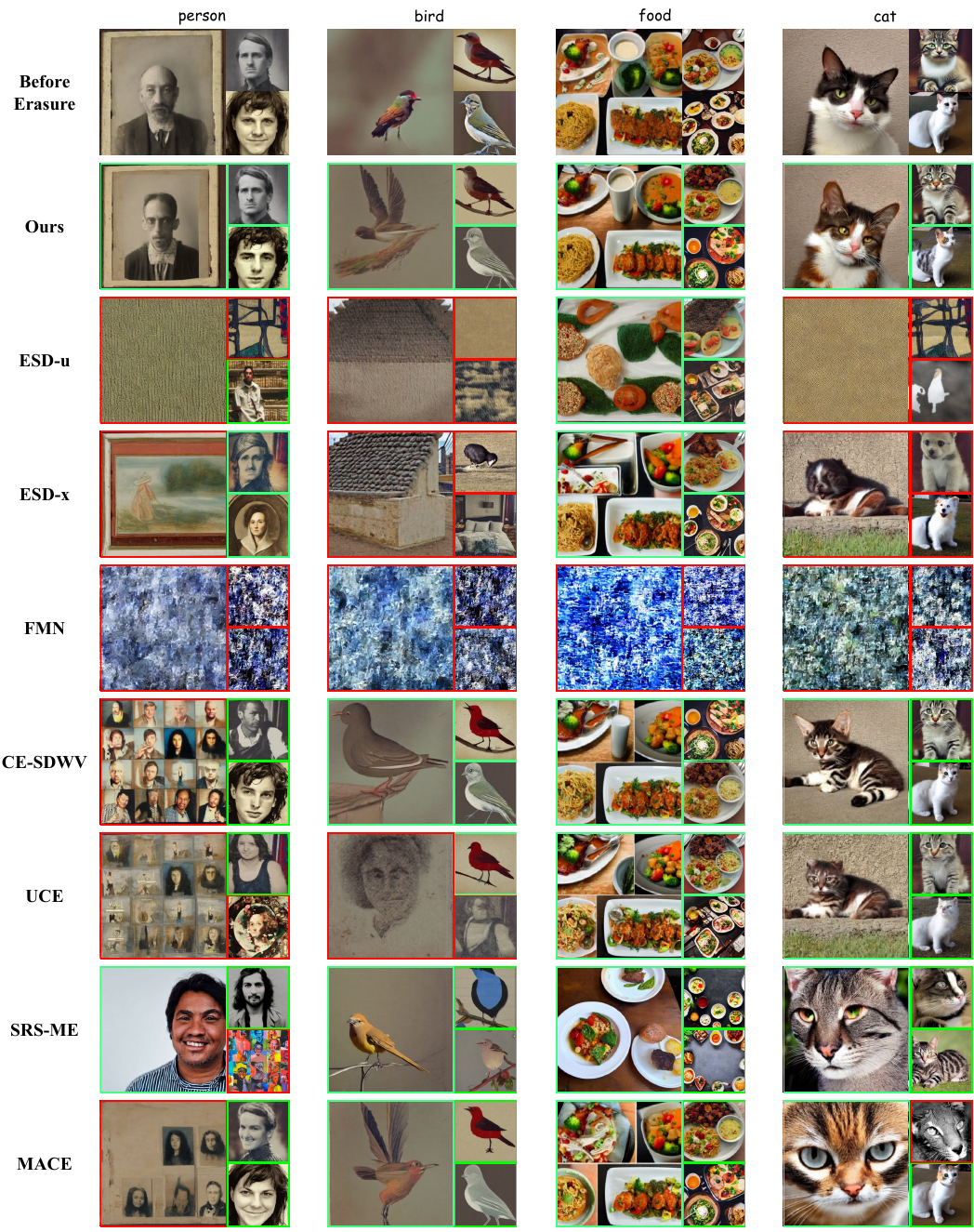}
   \vspace{-1em}
   \caption{Qualitative comparison of preserving supertype concepts.}
   \label{fig:appendix_preserve_supertype}
\end{figure*}



\end{document}